\pdfoutput=1

\documentclass[11pt]{article}

\usepackage[final]{acl}

\usepackage{times}
\usepackage{latexsym}

\usepackage[T1]{fontenc}

\usepackage[utf8]{inputenc}

\usepackage{microtype}

\usepackage{inconsolata}

\usepackage{graphicx}

\usepackage{hyperref}
\usepackage{booktabs}
\usepackage{adjustbox}
\usepackage{wrapfig}
\usepackage{amssymb}
\usepackage{color,xcolor,colortbl}
\usepackage{enumitem}
\usepackage{soul}
\usepackage{amsmath}
\usepackage{url}
\usepackage{algorithm}
\usepackage[noend]{algpseudocode}
\usepackage{booktabs}
\usepackage{bm,bbm}
\usepackage{mathtools}
\usepackage{array}
\usepackage{multirow}
\usepackage{subcaption}
\usepackage{pbox}
\usepackage{pifont}
\usepackage{arydshln}
\usepackage{dblfloatfix}
\usepackage{relsize}
\usepackage{setspace}
\usepackage[scaled=.8]{beramono}
\usepackage{tcolorbox}
\usepackage{listings}
\newcommand\numberthis{\addtocounter{equation}{1}\tag{\theequation}}

\newenvironment{fontppl}{\fontfamily{ppl}\selectfont}{\par} 

\newtcolorbox{promptbox}[2][]{
  floatplacement={#2},
  colframe=dark,colback=light!30!white,
  fonttitle=\small\ttfamily,
  fontupper=\small\ttfamily,
  title=#2,
  boxrule=0.5mm, 
  halign=flush left,
}

\definecolor{dark}{HTML}{064a6c}
\definecolor{light}{HTML}{efede1}

\title{STRUX: An LLM for Decision-Making with Structured Explanations}

\author{Yiming Lu,$^1$ Yebowen Hu,$^2$ Hassan Foroosh,$^2$ Wei Jin,$^1$ Fei Liu$^1$ \\[0.5em]
$^1$Emory University\\
$^2$University of Central Florida\\[0.5em]
\texttt{\{yiming.lu, wei.jin, fei.liu\}@emory.edu} \\ \texttt{\{yebowen.hu, hassan.foroosh\}@ucf.edu}}

\begin{document}
\maketitle
\begin{abstract}

Countless decisions shape our daily lives, and it is paramount to understand the how and why behind these choices. In this paper, we introduce a new LLM decision-making framework called STRUX, which enhances LLM decision-making by providing structured explanations. These include favorable and adverse facts related to the decision, along with their respective strengths. STRUX begins by distilling lengthy information into a concise table of key facts. It then employs a series of self-reflection steps to determine which of these facts are pivotal, categorizing them as either favorable or adverse in relation to a specific decision. Lastly, we fine-tune an LLM to identify and prioritize these key facts to optimize decision-making. STRUX has been evaluated on the challenging task of forecasting stock investment decisions based on earnings call transcripts and demonstrated superior performance against strong baselines. It enhances decision transparency by allowing users to understand the impact of different factors, representing a meaningful step towards practical decision-making with LLMs.

\end{abstract}

\section{Motivation}

Decision-making is complex, as it requires the evaluation of various determinants that can influence outcomes~\cite{eigner2024determinantsllmassisteddecisionmaking}. This ability is crucial across multiple fields, ranging from healthcare, where decisions can determine patient health outcomes~\cite{lehman2022learningasklikephysician}, to finance, where investment choices can impact financial stability~\cite{keith-stent-2019-modeling,liu2023fingptdemocratizinginternetscaledata}. For LLMs to be effective, they must not only identify relevant facts but also weigh the favorable and unfavorable aspects to reach insightful conclusions. To date, it remains unclear whether LLMs can effectively balance multiple factors in complex scenarios to make rational decisions.

LLMs also produce lengthy, plain text explanations that can sometimes overwhelm users with too much information or ambiguity~\cite{vafa-etal-2021-rationales,alkhamissi-etal-2023-opt,SycophanticLLM,ye-etal-2023-complementary,wang-etal-2024-rescue}. As we increasingly rely on those LLMs for critical decision-making, it is important to prioritize transparency and accountability~\cite{ludan-etal-2023-explanation}. We propose structuring these explanations into a table format, where each fact is listed with a `strength level' that measures its influence on the decision-making process. This approach not only facilitates review and modification of various facts by humans, but also enhances the transparency of the decisions made.

Further, a significant advantage of LLMs is their ability to reason through complex scenarios, which can enhance the decision-making processes \citep{shinn2023reflexionlanguageagentsverbal,yao2023treethoughtsdeliberateproblem,zeng2024uncertaintyfragilemanipulatinguncertainty,band2024linguisticcalibrationlongformgenerations}. Notably, DeLLMa \cite{liu2024dellmaframeworkdecisionmaking} uses classical decision theory to help LLMs make decisions under uncertainty. It infers a utility function through prompting and optimizes the expected utility using Monte Carlo estimation. \citet{feng2024birdtrustworthybayesianinference} calculate decision probabilities using a Bayesian model and present results on datasets such as Common2Sense \citep{singh-etal-2021-com2sense} and PlaSma \citep{brahman2023plasmamakingsmalllanguage}. In contrast, our approach involves fine-tuning an LLM with domain-specific knowledge to ensure it prioritizes supporting facts accurately. Training instances are generated via a series of reflection steps, without relying on human annotations.

\begin{figure*}
    \centering
    \includegraphics[width=5.9in]{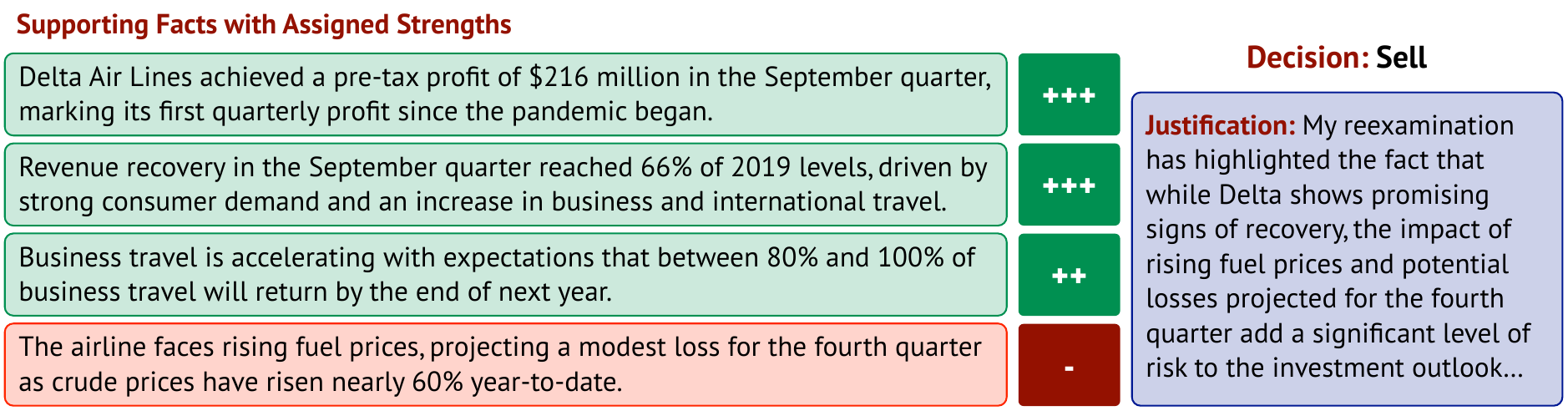}
    \vspace{-0.1in}
    \caption{STRUX's explanations consist of three components: \{supporting facts, a decision, and a brief justification\}. Supporting facts can include both positive (green) and negative (red) aspects, along with their strengths.}
    \label{fig:explanations}
    \vspace{-0.15in}
\end{figure*}

Our research explores the potential of using earnings call transcripts to forecast stock investment decisions~\cite{sawhney-etal-2020-voltage,medya2022exploratorystudystockprice,lopezlira2023chatgptforecaststockprice,ni2024harnessingearningsreportsstock}. Publicly traded companies in the U.S. are mandated by the Securities and Exchange Commission (SEC) to regularly report their financial performance, often through earnings calls. These calls include \emph{presentations from senior executives}, such as the CEO and CFO, followed by \emph{a Q\&A session} with financial analysts. The objective is to reassure investors about the company's management and strategy. With the rise of LLMs in financial services~\cite{zhu2021tatqa,sang-bao-2022-dialoguegat,cao2024risklabspredictingfinancialrisk,reddy2024docfinqa}, analyzing earnings call transcripts to guide stock investment decisions presents a promising opportunity to test the effectiveness of LLM-assisted decision-making.

\vspace{0.05in}
\noindent\textbf{Our research contributions include:} (a) we introduce STRUX, a novel framework designed to enhance the decision-making processes of LLMs. STRUX improves accuracy and transparency by meticulously constructing a fact table, analyzing these facts through a series of reflective steps, and fine-tuning the LLM to prioritize crucial information. (b) Our experiments demonstrate that STRUX surpasses strong baselines in forecasting stock investment decisions, proving its effectiveness. Its structured explanations further enhance decision transparency and represent a notable step towards practical decision-making with LLMs.\footnote{Our source code will be shared publicly upon acceptance.}

\section{The STRUX System}
\label{sec:strux}

STRUX is tasked with predicting a company's post-earnings stock trend to inform the investment decision. It is set to select the most relevant facts from a provided fact table, ensuring a balanced representation of positive and negative facts affecting the stock price. Each selected fact must then be evaluated for its potential impact on the stock's price movement. A ``+'' symbol indicates a positive impact, with the number of symbols varying from one (+) to three (+++) showing the increasing strength. Conversely, a ``-'' symbol denotes a negative impact, with one (-) to three (-{}-{}-) symbols reflecting the severity of the negative influence.

Our system then combines and analyzes all the selected facts to forecast the direction of the stock price movement. The outcomes include: Strongly Buy (SB), Buy (B), Hold (H), Sell (S), or Strongly Sell (SS). It also provides a justification elaborating on its rationale, focusing on the key facts that influence this decision. As illustrated in Figure~\ref{fig:explanations}, our \textbf{structured explanations} consist of three components: \{supporting facts, decision, and brief justification\}. Supporting facts can be both favorable and adverse, along with their respective strengths.

\subsection{Generating Structured Explanations Through Self-Reflection}

We create a fact table from each company's earnings call transcript to summarize key financial metrics, which are crucial for making informed investment decisions. Following~\citet{Koa_2024}, we input executive speeches from either the Prepared Remarks or Q\&A sessions into the LLM. Summaries are proportional in input length. Each speech from the Prepared Remarks is summarized into 3-5 key facts, while those from the Q\&A session are condensed into 1-3 key facts. The fact table was generated using OpenAI's \texttt{gpt-4o-mini-2024-07-18}; refer to the Appendix for the prompt. It distills essential information from a lengthy transcript, highlighting key aspects of a company's financials~\cite{cho-etal-2021-streamhover,cho-etal-2022-toward}.

\vspace{0.05in}
\noindent\textbf{Reflection.}\,\, We use a series of reflective steps to create training instances without requiring human annotations. This reflection was performed by GPT-4o-mini, aiming to help the model learn from its mistakes. When the model makes a poor investment decision, we notify it of the error and prompt it to identify any significant flaws in its fact selection, strength assignment, or reasoning processes. We also provide a list of previous incorrect decisions, including the reasons behind those decisions. Importantly, we ask the model to come up with a different decision from its previous ones \emph{without revealing the correct answer}. This approach allows us to observe the model's independent decision-making that emerges from reflection. Our prompt used for reflection can be found in the Appendix.

\vspace{0.05in}
\noindent\textbf{Demonstrations and Comparisons.}
Our `demonstrations' data contains training instances where output $\mathbf{y}$ has a correct decision post-reflection. We utilize this data to fine-tune \texttt{Llama3}, helping it prioritize relevant facts and make accurate decisions. The `comparisons' data consists of paired outputs, $\mathbf{y}$ and $\mathbf{y}^*$, where $\mathbf{y}^*$ is the output with the correct decision, and $\mathbf{y}$ is the prior model output in a series of reflections which has incorrect decision. These pairwise comparisons help train a reward model to favor outcomes that lead to correct decisions. Training instances that do not yield correct decisions after all reflections are excluded from demonstration or comparison data.

\subsection{Fine-tuning LLMs for Decision-Making}
\label{sec:supervised-finetuning}

\noindent\textbf{STRUX+SFT.}\,\, We start with the base LLM model, \texttt{Llama3-8b-Instruct}, and fine-tune it using our demonstrations data to develop the SFT model $p_\theta(\mathbf{y}|\mathbf{x})$. Specifically, the input $\mathbf{x}$ is a fact table created from an earnings call transcript, and the output $\mathbf{y}$ includes structured explanations that contain \{supporting facts, a decision, a brief justification\}. As illustrated in Equation~\ref{eq:sft}, the fine-tuning process aims to minimize the negative log-likelihood of the data. Here, $\mathbf{y}^* \sim \pi(\cdot | \mathbf{x})$ represents the demonstrations provided by \texttt{gpt-4o-mini-2024-07-18}, each of which contains the correct decision.
\begin{align*}
\mathcal{L}_{\text{SFT}}(\theta) = \mathop{-\mathbb{E}}_{\substack{\mathbf{x} \sim \mathcal{D}, \mathbf{y}^* \sim \pi(\cdot | \mathbf{x})}}\left[\, \log p_\theta(\mathbf{y}^*|\mathbf{x}) \,\right]
\numberthis\label{eq:sft}
\end{align*}

\begin{figure}[t]
    \centering
    \includegraphics[width=2.8in]{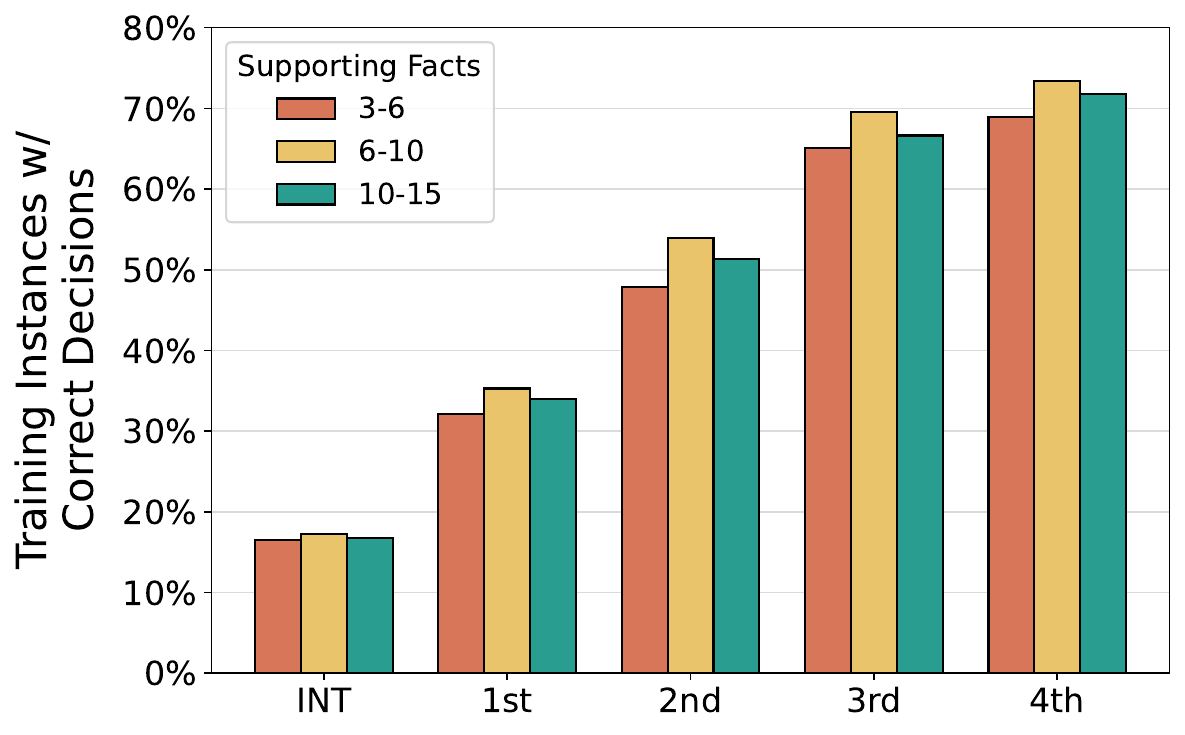}
    \vspace{-0.1in}
    \caption{Each iteration of self-reflection improves the accuracy of decision-making. We show the percentage of training instances that receive correct decisions after each iteration. Our STRUX model is instructed to select from three ranges of supporting facts: 3-6, 6-10, and 10-15. The selection of 6-10 supporting facts consistently yielded the highest accuracy.
    }
    \label{fig:num-facts}
    \vspace{-0.1in}
\end{figure}

\noindent\textbf{STRUX+RL.}\quad
In reinforcement learning, we start with a policy $p_{\theta'}(\mathbf{y} | \mathbf{x}) = p_{\theta}(\mathbf{y} | \mathbf{x})$ and fine-tune the policy $p_{\theta'}(\mathbf{y}|\mathbf{x})$ using a reward function $r_\phi(\mathbf{x}, \mathbf{y})$. We employ proximal policy optimization to optimize the expected reward. This process involves repeatedly choosing an instance from our training set, calculating the reward for the model's response with the reward function, then updating model parameters towards maximizing the reward. Following~\cite{ziegler2020finetuninglanguagemodelshuman}, we include a penalty $\beta \frac{p_{\theta'}(\mathbf{y}|\mathbf{x})}{p_\theta(\mathbf{y}|\mathbf{x})}$ to the reward to prevent $p_{\theta'}(\mathbf{y} | \mathbf{x})$ from diverging too far from $p_\theta(\mathbf{y}|\mathbf{x})$ where the learned reward $r_\phi(\mathbf{x}, \mathbf{y})$ is valid; $\beta$ is set to 0.2 in our study. 
\begin{align*}
\mathcal{L}_{\text{RL}}(\theta') = \mathop{-\mathbb{E}}_{\substack{\mathbf{x} \sim \mathcal{D}, \\\mathbf{y} \sim p_{\theta'}(\cdot | \mathbf{x})}} \left[r_\phi(\mathbf{x}, \mathbf{y}) - \beta \frac{p_{\theta'}(\mathbf{y}|\mathbf{x})}{p_\theta(\mathbf{y}|\mathbf{x})} \right]
\end{align*}
The reward function $r_\phi(\mathbf{x}, \mathbf{y})$ is trained using `comparisons' data. For every input $\mathbf{x}$, a response with the correct decision $\mathbf{y}^*$ is paired with $\mathbf{y}$, corresponding to the incorrect response \emph{prior to a successful reflection}. Below, $\sigma(r_\phi( \mathbf{x}, \mathbf{y}^*) - r_\phi(\mathbf{x}, \mathbf{y}))$ represents the probability that $\mathbf{y}^*$ is preferred over $\mathbf{y}$, denoted by $p(\mathbf{y}^* \succ \mathbf{y})$. We implement the reward $r_\phi(\mathbf{x}, \mathbf{y})$ as a linear function of the final embedding from the SFT model, and use this reward model to guide the policy learning during RL.
{\medmuskip=1mu
\thinmuskip=1mu
\thickmuskip=1mu
\nulldelimiterspace=1pt
\scriptspace=1pt
\begin{align*}
\mathcal{L}_{\text{RM}}(\phi) = \mathop{-\mathbb{E}}_{\substack{\mathbf{x} \sim \mathcal{D}, \\\mathbf{y}, \mathbf{y^*} \sim \pi(\cdot | \mathbf{x})}} \left[ \log \sigma(r_\phi( \mathbf{x}, \mathbf{y}^*) - r_\phi(\mathbf{x}, \mathbf{y})) \right]
\end{align*}}

\begin{table}[t]
    \setlength{\tabcolsep}{3.2pt}
    \renewcommand{\arraystretch}{1.1}
    \centering
    \begin{small}
    \begin{tabular}{lrrrrr}
    \textbf{System} & \textbf{Recall} & \textbf{Prec} & \multicolumn{1}{c}{\textbf{F$_1$}} & & \textbf{Accu.}\\
    \toprule
    Llama3-8b (\emph{Fact Table}) & 17.36 & 13.67 & 12.26 & & 16.70 \\
    GPT-4o-mini (\emph{Full Trans}) & 21.05 & 12.01 & 10.12 & & 17.21 \\
    GPT-4o-mini (\emph{Fact Table}) & 21.81 & 17.61 & 13.31 & & 20.27 \\
    DeLLMa {\scriptsize\cite{liu2024dellmaframeworkdecisionmaking}} & 38.30 & 23.14 & 16.68 & & 22.35 \\
    (Ours) STRUX+SFT & 19.15 & 15.55 & 16.54 & & 23.34 \\
    (Ours) STRUX+RL & 23.03 & 19.34 & \textbf{19.80} & & \textbf{25.55} \\
    \bottomrule
    \end{tabular}
    \end{small}
    \vspace{-0.1in}
    \caption{Our STRUX system outperforms strong benchmarks in making stock investment decisions. We present macro-averaged precision, recall, F-scores, accuracy for the test set. LLMs evaluated are: \texttt{Llama3-8b-Instruct} and \texttt{gpt-4o-mini-2024-07-18}.
    }
    \label{tab:test-set-results}
    \vspace{-0.2in}
\end{table}

\begin{table*}
    \setlength{\tabcolsep}{5pt}
    \renewcommand{\arraystretch}{1.1}
    \centering
    \begin{small}
    \begin{tabular}{lllllll}
    \multicolumn{3}{c}{\textbf{Frequent Paths Leading to \textcolor{blue}{Correct Decisions}}} & & \multicolumn{3}{c}{\textbf{Frequent Paths Leading to \textcolor{red}{Incorrect Decisions}}}\\
    \cmidrule[1pt]{1-3} \cmidrule[1pt]{5-7}
    B$\rightarrow$H (10.1\%) & \multicolumn{2}{l}{\quad\quad\quad\quad B$\rightarrow$H$\rightarrow$S$\rightarrow$SB (2.8\%)} & & B$\rightarrow$\textcolor{red}{H}$\rightarrow$SB$\rightarrow$S$\rightarrow$\textcolor{red}{H} (2.9\%) & & B$\rightarrow$H$\rightarrow$\textcolor{red}{S}$\rightarrow$SB$\rightarrow$\textcolor{red}{S} (1.5\%) \\
    B$\rightarrow$H$\rightarrow$SB (9.0\%) & & \quad\quad\, B$\rightarrow$S (2.5\%) & & B$\rightarrow$S$\rightarrow$\textcolor{red}{H}$\rightarrow$SS$\rightarrow$\textcolor{red}{H} (2.1\%) & & \textcolor{red}{B}$\rightarrow$S$\rightarrow$H$\rightarrow$SS$\rightarrow$\textcolor{red}{B} (1.4\%) \\
    \multicolumn{2}{l}{B$\rightarrow$H$\rightarrow$SB$\rightarrow$S (4.7\%)} & \quad\quad\, SB$\rightarrow$H (2.2\%) & & \textcolor{red}{B}$\rightarrow$H$\rightarrow$SB$\rightarrow$S$\rightarrow$\textcolor{red}{B} (2.0\%) & & SB$\rightarrow$H$\rightarrow$\textcolor{red}{B}$\rightarrow$S$\rightarrow$\textcolor{red}{B} (1.1\%) \\
    \bottomrule
    \end{tabular}
    \end{small}
    \vspace{-0.1in}
    \caption{The most common decision paths during reflection and their percentages in the training data. 
    }
    \vspace{-0.15in}
    \label{tab:paths}
\end{table*}

\vspace{-0.1in}
\section{Earnings Call Transcripts}
\label{sec:data}

Our dataset includes 11,950 quarterly earnings call transcripts from the \href{https://www.fool.com}{Motley Fool} website, collected by~\citet{hu2024defineenhancingllmdecisionmaking}, covering the period from 2017 to 2024. It contains transcripts from 869 companies listed on the NASDAQ 500 and S\&P 500, with an average of 10,187 tokens per transcript. Due to resource limits, we construct a balanced training set with 100 transcripts from each of the 11 financial sectors. Our test set consists of 587 transcripts from 2024, carefully chosen to ensure they were not part of the LLM pretraining, which has a cutoff up to December 2023. Our study focuses on the textual information of these transcripts and excludes acoustic features. The ground-truth investment decisions are based on a stock's performance 30 days post-earnings; they are categorized as Strongly Buy, Buy, Hold, Sell, or Strongly Sell.

\begin{table}[t]
    \setlength{\tabcolsep}{2pt}
    \renewcommand{\arraystretch}{1.1}
    \centering
    \begin{small}
    \begin{tabular}{lr}
    \toprule
    Total Number of Facts Per Transcript & 39.92 \\
    Num of Supporting Facts Per Transcript & 9.11\\
    Num of Favorable Supporting Facts & 8.01\\
    Favorable Facts with Strengths 1 to 3 & 1.00 / 4.53 / 2.48\\
    Number of Adverse Supporting Facts & 1.10 \\
    Adverse Facts with Strengths 1 to 3 & 0.58 / 0.29 / 0.23\\
    \bottomrule
    \end{tabular}
    \end{small}
    \vspace{-0.1in}
    \caption{Statistics of supporting facts. 
    }
    \label{tab:supporting-facts}
    \vspace{-0.2in}
\end{table}

\section{Experimental Results}
\label{sec:results}

We evaluated our STRUX against strong baselines for forecasting stock investment decisions. This includes DeLLMa~\cite{liu2024dellmaframeworkdecisionmaking}, which incorporates uncertainty into LLM decision-making using classical decision theory and has been tested on tasks such as agriculture planning and finance. Additionally, we tested \texttt{gpt-4o-mini-2024-07-18} and \texttt{Llama3-8b-Instruct} by providing either \emph{full transcripts} or concise \emph{fact tables} to elicit investment decisions; see Appendix for the prompt.

\vspace{0.05in}
\noindent\textbf{System Comparisons.}\quad Table~\ref{tab:test-set-results} shows the macro-averaged precision, recall, F-scores, and accuracy for the test set. STRUX outperforms strong baselines in accuracy and F-scores for stock investment decisions. Our findings indicate that adding reinforcement learning (STRUX+RL) leads to stronger performance compared to using the SFT method alone. We also find that direct prompting methods, e.g., GPT-4o-mini with Fact Table, tend to produce overly positive outcomes, often failing to suggest Strong Sell or Sell decisions. This bias can be traced back to the optimistic financial descriptions by company executives, and without fine-tuning, it leads LLMs to display a bias toward bullish predictions. It is also worth mentioning that our test set has an imbalanced label distribution. A random baseline achieves an accuracy of 19.11\%, and our STRUX+RL model shows a notable improvement, reaching an accuracy of 25.55\%.\footnote{We observe that OpenAI's \texttt{o1-mini-2024-09-12}, which generates a detailed internal thought process, only achieves a 16\% accuracy on this task, possibly due to overthinking.}

\vspace{0.05in}
\noindent\textbf{Supporting Facts.} We analyzed the supporting facts identified by the model in cases of correct decisions after reflections. Statistics are presented in Table~\ref{tab:supporting-facts}. Each transcript is distilled into a table of about 40 facts, from which the model selects 9. The selection is predominantly positive, with 8 positive and 1 negative fact; about half of the negative fact has an impact strength of 2–3. This indicates that adding expert knowledge on potential negative factors such as financial risks could make the fact tables more comprehensive. Figure~\ref{fig:num-facts} illustrates our experiment in which the model selects supporting facts from three ranges during self-reflection: 3-6, 6-10, and 10-15. We found that selecting 6-10 facts consistently yielded the highest performance.

\begin{figure}[t]
    \centering
    \includegraphics[width=2.8in]{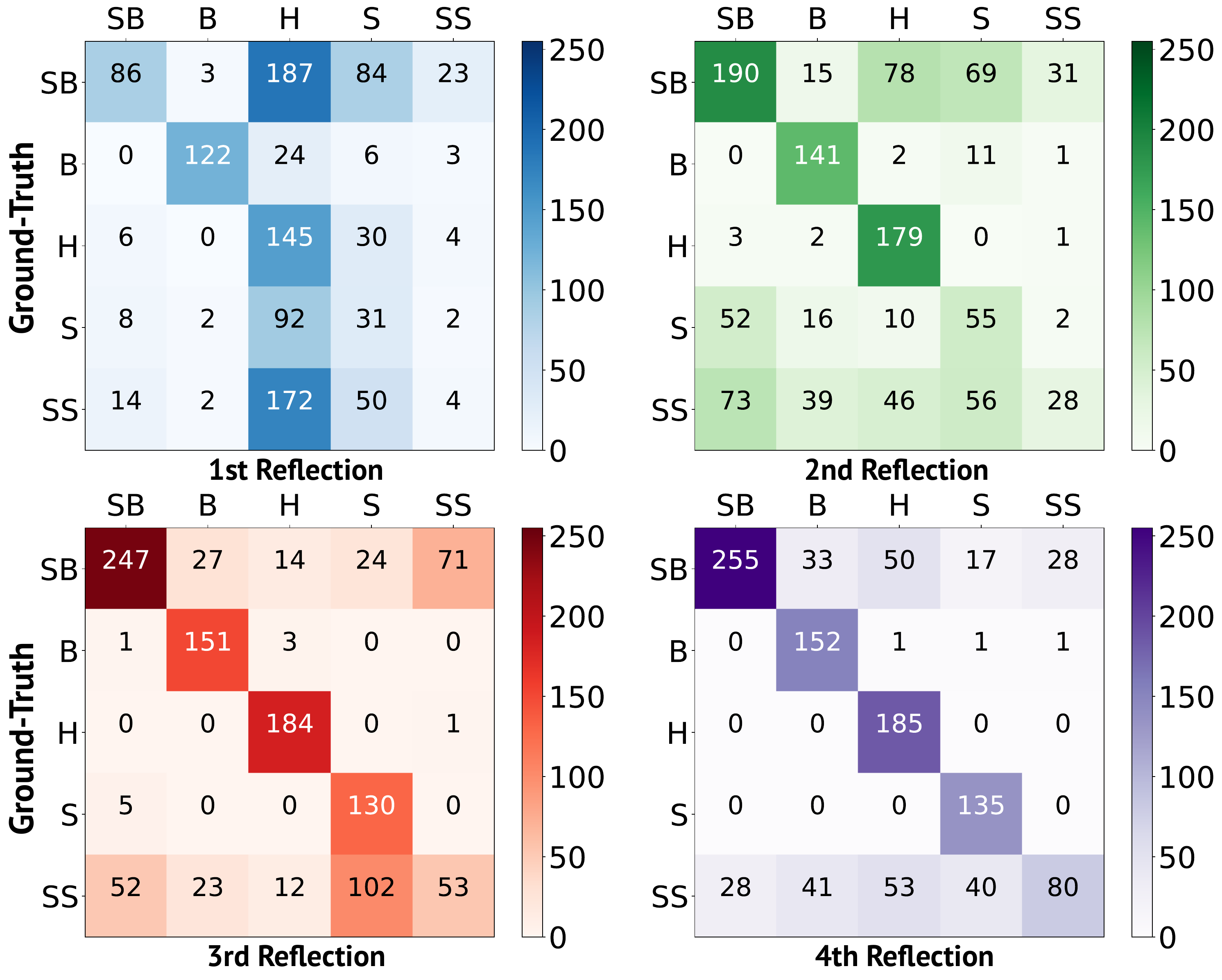}
    \vspace{-0.05in}
    \caption{Confusion matrix after each reflection. 
    }
    \label{fig:confusion-matrix}
    \vspace{-0.2in}
\end{figure}

\vspace{0.05in}
\noindent\textbf{Decision Paths.}\,\,\, STRUX performed 4 rounds of self-reflection, because there are 5 ground-truth decisions. Figure~\ref{fig:confusion-matrix} presents the confusion matrices, with each round of reflection improving the model's accuracy. The model initially favored `Hold' as a conservative decision. After two rounds of reflection, it began to predict decisions more accurately. Ultimately, the errors arise from the model's reluctance to recommend `Strong Sell' likely due to the postive language in executive speeches. 

Table~\ref{tab:paths} shows \emph{common decision paths} during reflection. Interestingly, reflection can lead to abrupt decision changes, such as a direct jump from Buy to Strong Sell, instead of gradual shifts (e.g., Buy $\rightarrow$ Hold $\rightarrow$ Sell). Moreover, reflection does not always yield perfect outcomes; the model can repeat decisions from previous cycles despite being instructed not to. These observations suggest that guardrails for self-reflection may help stabilize the decision-making process and prevent radical changes.

\section{Conclusion}
\label{sec:conclusion}

STRUX marks a notable step in using LLMs for decision-making. It integrates structured explanations into the decision-making process through a series of reflective steps. STRUX not only leads to higher accuracy but also improves the transparency of LLM decisions, making it a valuable tool for complex decision-making scenarios.

\section{Limitations}
\label{sec:limitations}

STRUX represents a significant advancement in using LLMs for decision-making, particularly in financial contexts. However, it's crucial to refine its fact extraction capabilities, as inaccuracies in data selection can impact decision quality. Additionally, predicting stock movements is inherently complex and influenced by various external factors like data quality and market nuances. Users are advised to carefully consider these aspects to maximize STRUX’s effectiveness and accuracy in real-world applications. With ongoing enhancements, STRUX has the potential to revolutionize decision-making across diverse sectors.

\bibliography{main}

\appendix

\section{Implementation Details}
\label{sec:hyperparameters} 

For STRUX+SFT, we fine-tune the system for three epochs with a learning rate of 1e-5, adjusted using a cosine scheduler. A warm-up ratio of 0.1 is set to ease the model into training, and we use the Adam optimizer configured with betas=(0.9, 0.999) and epsilon=1e-08. Our Reward Model (RM) also runs for three epochs, using a learning rate of 1e-4. It shares the same cosine scheduler and warm-up approach. For our STRUX+RL using Proximal Policy Optimization (PPO), the training lasts two epochs with the learning rate set to 1e-5. 

Our summarizer is instructed to focus on significant details that could impact the stock price, including \emph{financial performance, future outlooks and guidance, strategic decisions, company direction, market trends, competitive positioning, etc}. It also incorporates three historical financial metrics: \emph{earnings per share (EPS), revenue trends, and historical stock price}, gathered from \href{https://www.alphavantage.co}{Alpha Advantage}. These metrics are classified into three categories: `Bullish' (indicating strong financial health), `Stable' (showing steady metrics), and `Bearish' (suggesting investor pessimism). We focus on speeches from company executives and omit input from organizers and analysts.

\begin{figure*}[htbp]
\centering
\begin{footnotesize}
\begin{minipage}{\textwidth}

\begin{promptbox}{\fontppl Generating a Fact Table from an Earnings Call Transcript}

You have been given an executive's speech from an earnings call transcript. This could be from the Prepared Remarks segment or from responses given during the Q\&A session. Your task is to summarize the essential details related to \{company-ticker\} stock.\\[1em]

1. Keep your summary concise, with no more than \{number-of-facts\} key facts.\\
2. Focus on significant details that could impact the stock price, including financial performance, future outlooks and guidance, strategic decisions and company direction, market trends and competitive positioning, introductions of new products or services, and responses to industry challenges and opportunities.\\
3. Present these facts clearly without using any numbering or special formatting.\\
4. Make sure your summary remains factual and based solely on the content of the transcript.\\[1em]

**Examples:**\\

**Example 1 (Prepared Remarks):**\\[1em]

Earnings call transcript:

\begin{lstlisting}
"name": "John Smith, CEO",
"speech": [
    "Thank you, everyone, for joining us today. I'm pleased to report that our Q4 results exceeded expectations, with revenue growing 15\% year-over-year to \$2.5 billion. This growth was primarily driven by strong performance in our cloud services division, which saw a 30\% increase in revenue.",
    "However, we faced some challenges in our hardware segment, where revenue declined by 5\% due to supply chain disruptions. We're actively working to mitigate these issues and expect improvements in the coming quarters.",
    "Looking ahead, we're excited about the launch of our new AI-powered platform next month, which we believe will open up significant opportunities in the enterprise market. We're also continuing to invest heavily in R\&D, with a focus on sustainable technologies that we believe will drive long-term growth.",
    "In terms of guidance, we're projecting revenue growth of 10-12\% for the next quarter, which is slightly below analyst estimates due to ongoing macroeconomic uncertainties."
]
\end{lstlisting}

Facts:
\begin{lstlisting}
    Company reported Q4 revenue of $2.5 billion, a 15% year-over-year increase, exceeding expectations.
    Cloud services division saw a 30% increase in revenue, driving overall growth.
    Hardware segment revenue declined by 5% due to supply chain disruptions.
    New AI-powered platform launching next month expected to create significant opportunities in the enterprise market.
    Company is investing heavily in R&D, focusing on sustainable technologies for long-term growth.
    Guidance for next quarter projects 10-12% revenue growth, slightly below analyst estimates.
\end{lstlisting}

**Example 2 (Q\&A Session):**\\[1em]

Earnings call transcript:

\begin{lstlisting}
"name": "John Smith, CEO",
"speech": [
    "The 5\% decline in our hardware segment was primarily due to semiconductor shortages affecting our production capacity. We've already secured new suppliers and expect to resolve most of these issues by the end of next quarter. In fact, we anticipate returning to growth in this segment by Q3."
]
\end{lstlisting}

Facts:
\begin{lstlisting}
    Hardware segment declined 5% due to semiconductor shortages; new suppliers secured, issues expected to be resolved by next quarter end.
    Anticipate returning to growth in hardware segment by Q3.
\end{lstlisting}

Earnings call transcript:
\{earnings-call-transcript\}\\[0.5em]

Facts:"""

\end{promptbox}

\end{minipage}
\end{footnotesize}
\vspace{-0.1in}
\caption{We input executive speeches from the Prepared Remarks or Q\&A sessions into the LLM. Summaries are proportional in input length. Each speech from the Prepared Remarks is summarized into 3-5 key facts, while those from the Q\&A session are condensed into 1-3 key facts. Fact tables are generated using \texttt{gpt-4o-mini-2024-07-18}.}
\end{figure*}

\begin{figure*}[htbp]
\centering
\begin{footnotesize}
\begin{minipage}{\textwidth}

\begin{promptbox}{\fontppl Predicting a Company's Post-Earnings Stock Trend to Inform the Investment Decision}

Your task is to make an investment decision by predicting the post-earnings stock movement trend for \{company-ticker\} over a 30-day period. Use the provided fact table and follow these steps:\\[0.5em]

\begin{lstlisting}
1. Choose 6-10 of the most relevant facts from the table. Make sure there is a balance between positive and negative facts.

2. Each selected fact needs to be assessed for its likely impact on the stock's price:

  - Use a '+' symbol to denote a positive impact. The number of '+' symbols can vary from one ('+') to three ('+++') depending on the increasing strength of the positive impact.
   
  - Use a '-' symbol to denote a negative impact. Similarly, the number of '-' signs can range from one ('-') to three ('---') based on the severity of the negative impact.

3. Prioritize facts that could influence the stock price over the long term.

4. Evaluate the facts based on both the quantitative (impact strengths) and qualitative (relevance and importance) aspects of each fact.

5. Combine and analyze all the selected facts to predict the likely direction of the stock price movement.

Your response must be formatted as follows:

Selected Facts with Assigned Strength:
   - [Fact 1] | [Content]: [Assigned Strength]
   - [Fact 2] | [Content]: [Assigned Strength]
   ...
   (Include between 6-10 facts with their assigned strengths)

Decision: [Choose one: Strongly Buy, Buy, Hold, Sell, Strongly Sell. Please note that no other responses will be considered valid.]

Justification: [Provide a concise paragraph summarizing your reasoning, focusing on key facts that influence your decision.]

Fact Table: {fact-table}
\end{lstlisting}

\end{promptbox}

\end{minipage}
\end{footnotesize}
\vspace{-0.1in}
\caption{STRUX is tasked with predicting a company's post-earnings stock trend to inform the investment decision. It is set to select the most relevant facts from a provided fact table, ensuring a balanced representation of positive and negative facts affecting the stock price. Each selected fact is evaluated for its potential impact on the stock's price movement. A ``+'' symbol indicates a positive impact, with the number of symbols varying from one (+) to three (+++) showing the increasing strength. Conversely, a ``-'' symbol denotes a negative impact, with one (-) to three (-{}-{}-) symbols reflecting the severity of the negative influence. The system then analyzes all the selected facts to forecast the direction of the stock price movement. The outcomes include: Strongly Buy (SB), Buy (B), Hold (H), Sell (S), or Strongly Sell (SS). It also provides a justification elaborating on its rationale, focusing on the key facts that influence this decision. Additionally, we tested \texttt{gpt-4o-mini-2024-07-18} and \texttt{Llama3-8b-Instruct} using this prompt by providing either \emph{full transcripts} or concise \emph{fact tables} to elicit investment decisions.}
\end{figure*}

\begin{figure*}[htbp]
\centering
\begin{footnotesize}
\begin{minipage}{\textwidth}

\begin{promptbox}{\fontppl Reflecting on Past Errors to Enhance the Model's Decision-Making Abilities}

You are an advanced reasoning agent capable of enhancing your capabilities through self-reflection. In a previous task, you analyzed a fact table related to a specific stock. You selected various facts from the table, assigned impacts and strengths to them, and formulated a stock investment decision along with supporting justifications. Unfortunately, your assessments led to an incorrect stock investment decision.\\[1em]

Your current task is to critically review your prior efforts. You must reexamine the original fact table, the facts you previously selected, the strengths you assigned to each, and the reasoning behind your conclusions. It is essential to identify significant flaws in your selection of facts, the assignment of their strengths, or in the reasoning process you employed.\\[1em]

You must adhere to the following format in your analysis. Any deviation from this format will render it invalid. Your new stock investment decision should differ from all previous ones and should be derived exclusively from a detailed analysis of the provided facts, without relying on any pre-existing patterns.\\[0.5em]

\begin{lstlisting}
========
INPUT:

Fact Table: 
[The full fact table will be provided here]

Previous Incorrect Outputs: 
[A list of previously incorrect outputs will be included here, containing selected facts, their assessed strengths, decisions, and the justifications provided for them.]

OUTPUT:

Selected Facts with Assigned Strength:
- Fact [number] | [Content]: [Assigned Strength]
- [This pattern will continue for each of the selected facts, ensuring that 6-10 facts are chosen.]

Decision:
[Your new decision, which must be different from all previous decisions, will be one of the following: Strong Buy, Buy, Hold, Sell, Strong Sell.]

Justification:
[Provide a clear explanation for your updated changes and new decision in a single paragraph. Emphasize how your analysis of the facts led you to a different decision from previous outputs, and how you have addressed any errors found in prior assessments.]

========
INPUT: 

Fact Table:
{fact-table}

Previous Incorrect Outputs: The following list includes outputs from previous trials. This includes decisions that were incorrect, potentially incorrect facts that were selected, and inaccurately assigned strengths.
{previous-incorrect-outputs}

OUTPUT:
\end{lstlisting}

\end{promptbox}

\end{minipage}
\end{footnotesize}
\vspace{-0.1in}
\caption{We use a series of reflective steps to create training instances without requiring human annotations. This reflection was performed by \texttt{gpt-4o-mini-2024-07-18}, aiming to help the model learn from its mistakes. When the model makes a poor investment decision, we notify it of the error and prompt it to identify any significant flaws in its fact selection, strength assignment, or reasoning processes. We also provide a list of previous incorrect decisions, including the reasons behind those decisions. Importantly, we ask the model to come up with a different decision from its previous ones \emph{without revealing the correct answer}. This approach allows us to observe the model's independent decision-making that emerges from reflection.}
\end{figure*}

\begin{figure*}[htbp]
\centering
\begin{footnotesize}
\begin{minipage}{\textwidth}

\begin{promptbox}{}
\begin{lstlisting}
[Prepared Remarks:]

>> Operator

Good morning, everyone, and welcome to the Delta Air Lines September-quarter 2021 financial results conference call. My name is Jen, and I will be your coordinator. [Operator instructions] As a reminder, today's call is being recorded. I would now like to turn the conference over to Ms. Julie Stewart, vice president of investor relations. Please go ahead.

>> Julie Stewart -- Vice President of Investor Relations

Thank you, Jen. Good morning, everyone, and thanks for joining us for our September-quarter 2021 earnings call. Joining us from Atlanta today are CEO, Ed Bastian; our president, Glen Hauenstein; our CFO, Dan Janki. And Ed will open the call with an overview of Delta's performance and strategy.

Glen will provide an update on the revenue environment and our brand momentum, and Dan will discuss cost fleet and our balance sheet. Similar to last quarter's call, we've scheduled today's call for 90 minutes to make sure we have plenty of time for questions. [Operator instructions] After the analyst Q\&A, we will move to our media questions, after which, Ed will provide a brief closing statement. Today's discussion contains forward-looking statements that represent our beliefs or expectations about future events.

All forward-looking statements involve risks and uncertainties that could cause the actual results to differ materially from the forward-looking statements. Some of the factors that may cause such differences are described in Delta's SEC filings. We also discuss non-GAAP financial measures, and all results exclude special items unless otherwise noted. You can find a reconciliation of our non-GAAP measures on the Investor Relations page at ir.delta.com. And with that, I'll turn the call over to Ed.

>> Ed Bastian -- Chief Executive Officer

Well, thank you, Julie, and good morning, everyone. Appreciate you joining us this morning. The September quarter marked another important milestone in our recovery. We achieved our first quarterly profit since the start of the pandemic with a pre-tax result of $216 million and a pre-tax margin of nearly 3% despite still missing one-third of our revenue base compared to the same period in 2019... [omitted.]

[Questions & Answers:]

>> Operator

Thank you. And we'll go first to Jamie Baker with J.P. Morgan.

>> Jamie Baker -- J.P. Morgan -- Analyst

Hey. Good morning, everybody. First question goes potentially to Glen and Dan. So pre-COVID, I had asked Paul about the amount of time that it would typically take Delta to recalibrate the higher fuel prices.

I'm not staring at the transcript, but his estimate at the time was four to six months, which was an improvement from historic levels. So my question, I guess, for Glen is whether the booking curve is steep enough right now that you might actually be able to recapture the top line more quickly than that. And similarly, for Dan, whether there's anything we should be taking on the cost or operations side that could accelerate the process. I'm basically just trying to understand whether four to six months is still the right estimate for us to be using.

>> Glen Hauenstein -- President

Well, I would just comment, I think we're a bit in uncharted territory here as the recovery continues. And while I think it might be difficult in the very short run, despite the fact that the booking curve has moved in a bit, that I would estimate that, that four to six months is about right because we believe that demand and capacity will fall back into a very good equilibrium by next spring which would put you inside that window... [omitted.]

\end{lstlisting}

\end{promptbox}

\end{minipage}
\end{footnotesize}
\vspace{-0.1in}
\caption{An example of an earnings call transcript from Delta Air Lines (DAL) for Q3 2021.
}
\end{figure*}

\end{document}